\def\year{2022}\relax
\documentclass[letterpaper]{article} 
\usepackage{aaai22}  
\usepackage{times}  
\usepackage{helvet}  
\usepackage{courier}  
\usepackage[hyphens]{url}  
\usepackage{graphicx} 
\usepackage{natbib}  
\usepackage{caption} 
\DeclareCaptionStyle{ruled}{labelfont=normalfont,labelsep=colon,strut=off} 
\usepackage{algorithm}
\usepackage{algorithmic}

\usepackage{newfloat}
\usepackage{listings}
\floatstyle{ruled}
\newfloat{listing}{tb}{lst}{}
\floatname{listing}{Listing}
\title{FedFR: Joint Optimization Federated Framework for Generic and Personalized Face Recognition}
\author{
    Chih-Ting Liu\textsuperscript{\rm 1\footnote{Both authors contributed equally to this work.}},~Chien-Yi Wang\textsuperscript{\rm 2\footnotemark[1]},~Shao-Yi Chien\textsuperscript{\rm 1},~Shang-Hong Lai\textsuperscript{\rm 2}
}
\affiliations{
    \textsuperscript{\rm 1} Graduate Institute of Electronics and Engineering, National Taiwan University\\
    \textsuperscript{\rm 2} Microsoft AI R\&D Center, Taiwan\\
    jackieliu@media.ee.ntu.edu.tw, chiwa@microsoft.com, sychien@ntu.edu.tw, shlai@microsoft.com
}
\vspace{-2mm}

\usepackage{bibentry}
\usepackage{xcolor}

\usepackage{pifont}
\newcommand{\cmark}{\ding{51}}%
\newcommand{\xmark}{\ding{55}}%
\newcommand{\ra}[1]{\renewcommand{\arraystretch}{#1}}
\usepackage{multirow}
\usepackage{amsmath}
\usepackage{makecell}
\usepackage{comment}
\newif\ifsubmit
\submitfalse
\ifsubmit
\newcommand{\jerry}[1]{}

\newcommand{\ctliu}[1]{}

\else
\newcommand{\jerry}[1]{{\bf \textcolor{magenta}{Jerry: #1}}}

\newcommand{\ctliu}[1]{{\bf \textcolor{red}{Ctliu: #1}}}

\fi

\begin{document}

\maketitle

\begin{abstract}
Current state-of-the-art deep learning based face recognition (FR) models require a large number of face identities for central training. However, due to the growing privacy awareness, it is prohibited to access the face images on user devices to continually improve face recognition models. Federated Learning (FL) is a technique to address the privacy issue, which can collaboratively optimize the model without sharing the data between clients. In this work, we propose a FL based framework called FedFR to improve the generic face representation in a privacy-aware manner. Besides, the framework jointly optimizes personalized models for the corresponding clients via the proposed Decoupled Feature Customization module. The client-specific personalized model can serve the need of optimized face recognition experience for registered identities at the local device. To the best of our knowledge, we are the first to explore the personalized face recognition in FL setup. The proposed framework is validated to be superior to previous approaches on several generic and personalized face recognition benchmarks with diverse FL scenarios. The source codes and our proposed personalized FR benchmark under FL setup are available at \small \url{https://github.com/jackie840129/FedFR}.
\end{abstract}
\begin{figure}[t]
	\centering
    \includegraphics[width=0.9\linewidth]{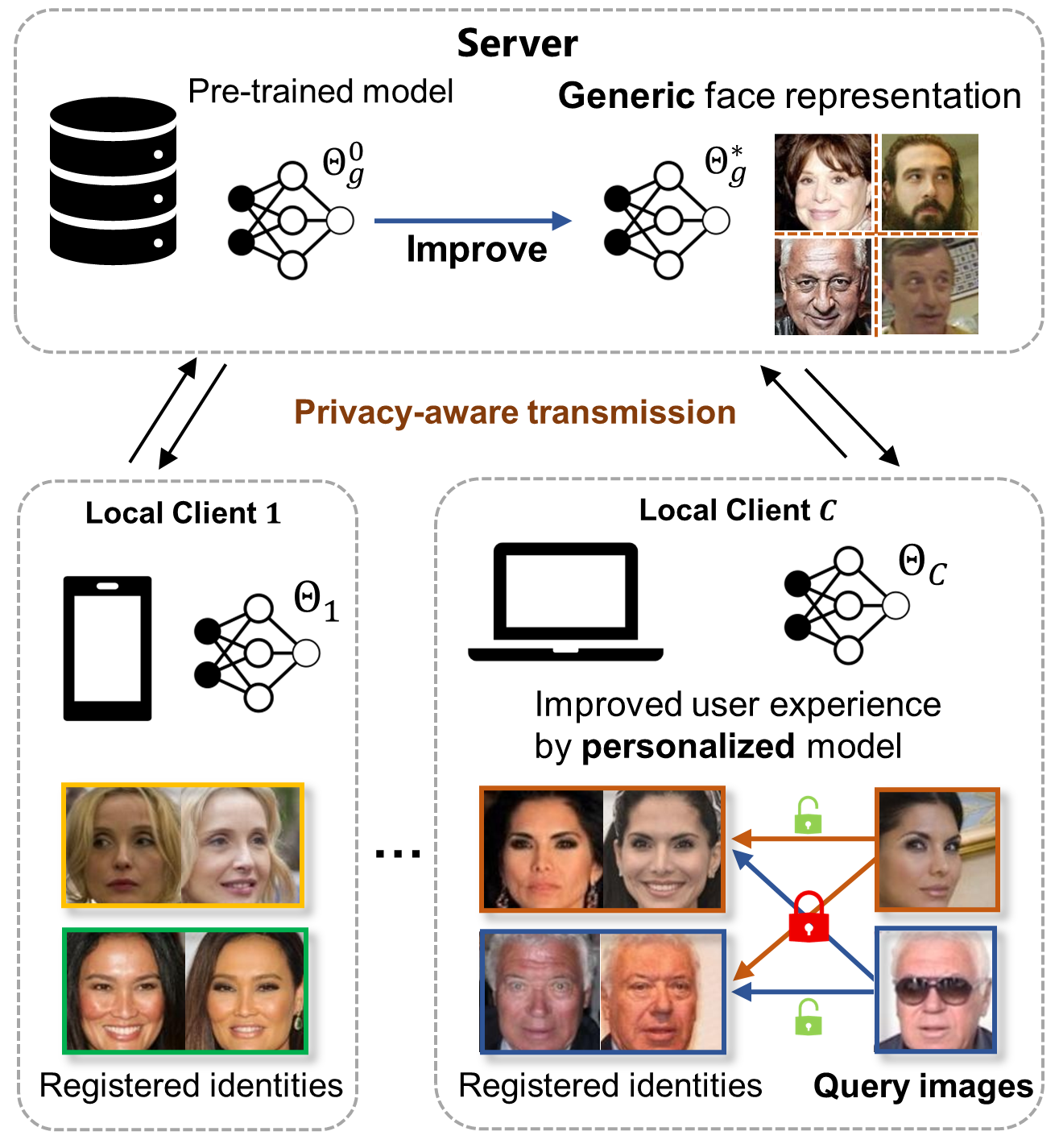}
    \caption{The Federated Learning (FL) setup for face recognition. Given a pre-trained face recognition model, we aim to simultaneously improve the generic face representation at the server, and produce an optimal personalized model for each client without transmitting private identities' images or features out of the local devices.}
    \label{fig:intro}
    \vspace{-3mm}
\end{figure}

\vspace{-2mm}
\section{Introduction}
Face recognition has been an active and vital topic among computer vision community for a long time. The state-of-the-art training frameworks formulate face recognition as a metric learning problem, and employ the large-scale identity classification as the proxy task to learn face features, which could discriminate between different identities robustly. Recently, the quick evolution of softmax-based loss functions for identity classification greatly promote the performance of face recognition. However, the training of face recognition model heavily relies on centralizing a huge amount of personal face images, which are usually not accessible due to the uprising privacy concern in many countries. Therefore, it is necessary to navigate the development of face recognition under the premise of privacy preservation.

Federated learning (FL) provides a distributed and privacy-aware framework to train models where multiple clients collaboratively learn without sharing their data with the central server or other clients. A classical FL method called FedAvg~\cite{mcmahan2017communication} aggregates and averages the gradients from local clients on the server, and transmit the updated model back to the clients for the next round of local optimization. In the past few years, there has been significant progress in FL~\cite{kairouz2019advances} on image classification task, which boosts the performance of aggregated global model under diverse FL scenarios. However, these approaches cannot be directly applied onto face recognition due to several critical reasons: 1) Face recognition is an open-set classification task, where training and testing identity classes are different. 2) The identity classes between local clients are different, which results in different model architectures in clients. 3) In a more practical setup for face recognition~\cite{aggarwal2021fedface}, the FL training starts from a publicly available face recognition model, rather than from scratch as in traditional FL.

In order to address these aforementioned issues, a recent work FedFace~\cite{aggarwal2021fedface} proposed an FL framework for face recognition model training in a privacy-aware manner. It tackles the challenging setup where each of the participating clients has face images of only one identity. It employs a mean feature initialization method for the local identity proxy and a spreadout regularizer~\cite{yu2020federated} at the server side to ensure that the identity proxies from the local clients are well separated. However, FedFace is limited as it only addressed a single scenario. In the real-world face recognition applications, local edge devices could be registered by multiple identities. Moreover, there exists a serious privacy concern in FedFace as it requires the local device to transmit the identity proxy to the server, which could violates the FL protocol~\cite{duong2020vec2face}. A concurrent work~\cite{meng2022improving} tries to mitigate this privacy concern through the Differential Privacy approach.  

To enable federated learning in more realistic face recognition settings, we propose a novel framework called \textbf{FedFR}, which could jointly improve generic and personalized face representations without breaking the privacy on clients. First, we leverage the globally shared dataset to regularize the training on local clients, as the local client only has much less identities than the pre-trained dataset. With the additional transmission of the shared class embedding matrix, it can effectively prevent the local model from over-fitting and also improve the generic representation at the server. Secondly, in order to reduce the computation overhead and improve the training efficiency, a novel hard negative sampling strategy is proposed to select the most critical data samples from the globally shared dataset. In addition, a contrastive loss applied on the local face representation during training could further restrict the local model drifting. Last but not least, we are interested in simultaneously optimizing the user experience on local clients, which is not explored in previous works. Although personalized FL~\cite{kulkarni2020survey} has been studied for a while, those methods are sub-optimal on the face recognition task. We propose a Decoupled Feature Customization (\textbf{DFC}) module, which consists of a feature transformation layer and one-vs-all binary classifiers. The module locally learns a customized feature space which is optimized for recognizing the registered identities at each client. 

We validate FedFR on IJB-C~\cite{maze2018iarpa} dataset for the generic recognition model performance under different FL scenarios. We also build the personalized face recognition evaluation protocol with MS-Celeb-1M~\cite{guo2016ms} dataset to validate the effectiveness of the proposed DFC module. Each technique in FedFR could substantially improve both generic and personalized face representations. Our main contributions are summarized as follows: 
\begin{itemize}
    \item We propose a novel joint optimization federated learning framework FedFR, which can effectively improve both generic and personalized face recognition models under different scenarios while strictly following the privacy constraints.
    \item Several training techniques (hard negative sampling, contrastive regularization) are proposed and tailored for the face recognition task, and these techniques can better bridge the gap between global and local representations.
    \item We propose the Decoupled Feature Customization (DFC) module, which is the key component to enable concurrent optimization of the personalized face recognition model. The proposed binary classification objectives are also effective for optimizing the performance on each client.
    \item Experimental results show that our proposed solution can consistently outperform previous approaches in several challeging generic and personalized FL benchmarks.
\end{itemize}
\vspace{-2mm}
\section{Related Work}
\noindent\textbf{Face Recognition.}
Recently, great progress has been achieved in face recognition with large-scale training data~\cite{cao2018vggface2, guo2016ms, zhu2021webface260m}, sophisticated network structures~\cite{schroff2015facenet, he2016deep} and advanced designs for softmax-based loss functions~\cite{wang2018cosface, deng2019arcface, sun2020circle}. However, these state-of-the-art methods are not directly applicable to the federated learning setting since they assume centralized data is available on a server. Without the access to private face images from local clients, the feature learning is prohibited as the model cannot compare features between different identities. In addition, how to leverage additional identities to improve the feature incrementally based on a pre-trained face recognition model was never discussed in previous works, as they always assumed to train the model from scratch. In our federated setup, we aim to improve a publicly available pre-trained face recognition model at the server from multiple clients in a collaborative manner, while keeping the private face images and identity features at the local clients.  

\noindent\textbf{Federated Learning.}
Federated Learning (FL)~\cite{li2019survey, kairouz2019advances, wang2021field} is a learning setup in machine learning which aims to learn a model over multiple disjoint clients while maintaining local data privacy. The most well-known and commonly used FL algorithm is FedAvg~\cite{mcmahan2017communication}, which learns a global model by averaging weight parameters across local models trained on private client datasets. Many recent works proposed to improve FedAvg from different perspectives: model convergence~\cite{haddadpour2019convergence, khaled2020tighter}, robustness~\cite{bonawitz2019towards}, communication~\cite{konevcny2016federated}, and non-IID clients~\cite{li2019convergence, li2021model}. 
Most of the previous computer vision related FL works only studied image classification tasks with small-scale datasets (e.g. MNIST, CIFAR-10). To the best of our knowledge, FedFace~\cite{aggarwal2021fedface} is the only one which addressed the face recognition model training in the federated setup. To enhance the pre-trained FR model, it applies the spreadout regularizer~\cite{yu2020federated} at the server side to ensure the identity proxies from clients are well separated. Our work differs in that we do not transmit identity prototypes as it could leak the private identity info from clients. Moreover, our work is scalable to different scenarios where each client contains more than one identity. 

\noindent\textbf{Personalized Federated Learning.}
Personalized FL~\cite{kulkarni2020survey} aims to learn a customized model to meet each client's objective. Instead of training a single ``general" model which is optimized for generic metric, this FL setup seeks to acknowledge the data heterogeneity among clients by constructing a ``personalized" model which fits each client's need. Many recent techniques~\cite{liang2020think, li2021fedbn, chen2021bridging} proposed to leverage multi-task learning (MTL)~\cite{zhang2017survey} methods to incorporate clients' task objectives into the FL framework. Another stream of approaches~\cite{chen2018federated, fallah2020personalized} employed meta-learning to learn a decent initial model that can be adapted to each client after some steps of local fine-tuning. Besides, ~\cite{yu2020salvaging} showed that conducting post-processing (e.g. fine-tuning) onto a generic FL model could achieve comparable results with other personalized methods. However, the latter two streams of approaches would require an additional stage for local adaptation. Our framework employs the MTL based approach which can optimize general and customized face recognition models simultaneously. 
\vspace{-2mm}
\section{Proposed Method}
In this work, we build a novel FL framework for the face recognition (FR) task. In the following, we will first establish the proposed FL setup for joint generic and personalized face recognition. Next, we introduce some preliminaries of our framework, which are some basic techniques popularly employed in FR and FL respectively. Then, we will describe technical details of the proposed FedFR solution. 

\begin{figure*}[t]
	\centering
    \includegraphics[width=0.93\textwidth]{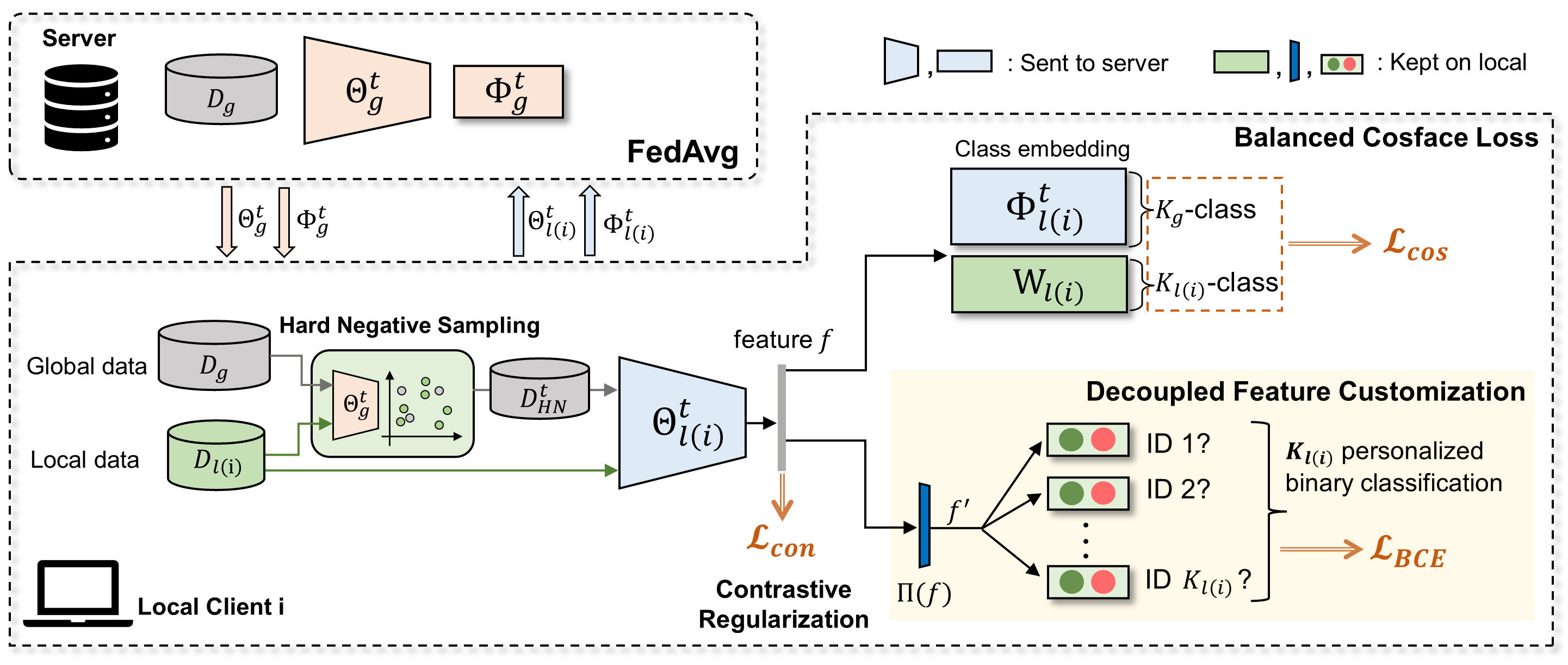}
    \caption{FedFR. We demonstrate the overall architecture of our method. For model on each client $i$, it will be optimized with balanced Cosface loss, a contrastive regularization and the binary cross entroy in our Decoupled Feature Customization branch.  After training, the backbone model $\Theta_{l(i)}$ and global class embedding $\Phi_{l(i)}$ will be uploaded for FedAvg.}
    \label{fig:big}
    \vspace{-3mm}
\end{figure*}

\vspace{-1mm}
\subsection{Problem Setup}
\label{sec:ps}
Face recognition systems are widely applied on local user devices. Typically, the deployed model is trained on a public dataset in advanced on a server. To continuously improve the generic face representation, the intuitive way is to collect the images stored in local devices (clients) and update the model trained with augmented data. However, as mentioned previously, due to privacy issues, it is prohibited to upload any identity-related information, such as the face images and its features. Federated learning (FL) provides a framework to train models where multiple clients collaboratively learn without sharing their data with the server or with other clients. Different from typical FL setting that learns the model from scratch, in face recognition, we target on \textbf{how to enhance the generic representation of pre-trained model by leveraging the data on clients under the privacy constraint}. Besides, we also focus on the optimized user experience. Although an improved generic model can implicitly achieve it, a client-specific personalized model optimized by local objectives could achieve optimal performance on the device. Thus, we jointly consider the situation that \textbf{whether we can obtain a personalized face model which is dedicated to recognize the registered identities on each client}. To the best of our knowledge, we are the first to explore the personalized FL setup in face recognition.
\vspace{-1mm}
\subsection{Preliminaries}
\label{sec:pre}
\paragraph{Face Recognition.} FR is an open-set problem, where the classes (identities) in training and testing are different. In the training phase, current FR methods are typically based on an identity classification objective, where the model embeds an input image into a high-dimensional representation and generates the class logits by computing the similarity between the input feature and all class embeddings (proxies). Then a softmax cross-entropy loss will be adopted to supervise the model. In our setting, the pre-trained generic face model is trained with the commonly used Cosface loss~\cite{wang2018cosface}, which adopts an additive margin softmax. Formally, given the face embedding model $\Theta$ and an input image $x$ with $y$-th class, we can obtain its deep feature $f=\Theta(x) \in \mathcal{R}^{d}$. There is also a class embedding matrix $\Phi \in \mathcal{R}^{d\times K}$, where $K$ is the total number of classes and the $j$-th column $\Phi_j$ means the learned proxy of $j$-th class. Following Cosface loss, the original $j$-th logit ($\Phi_j \cdot f+b$) will be simplified by ignoring the bias $b$ and normalizing the $\|f\|$ and $\|\Phi_j\|$ to $1$, which is just the cosine similarity $\cos\theta_j$. Last, the additive margin softmax cross-entropy loss for $x$ will be computed as follows:
\begin{equation}
    \mathcal{L}_{cos} = - \log \frac{e^{s(\cos\theta_{y}-m)}}{e^{s(\cos\theta_{y}-m)} + \sum_{j\neq y}^{K}e^{s\cos\theta_{j}}},
\end{equation}
where $s$ and $m$ are the scaling constant and the additive margin, respectively. During the testing stage, the learned face embedding model $\Theta$ will embed the query face image into a $d$-dim face feature, and the system would compare the cosine distance between the query feature and pre-registered features for identity authentication.

\paragraph{Federated Learning.} In our FL setup, we consider $C$ local client nodes and one central server with the face recognition model $\Theta_{g}^{0}$ pre-trained on a publicly available large dataset $D_g$, which has $N_g$ images from $K_g$ identities. Each local client $i$ is initialized with $\Theta^0_{l(i)}=\Theta^0_g$ and registered with $N_{l(i)}$ images from $K_{l(i)}$ identities, which is much smaller than the public one. Our objective is to simultaneously improve the model $\Theta_g$ for generic face representation and optimize each $\Theta_{l(i)}$ for personalized client customization under the privacy constraints. We adopt the most commonly used FL algorithm, \textbf{FedAvg}~\cite{mcmahan2017communication}, as our baseline method. Due to the mutual exclusive classes between local clients, we follow previous FL works~\cite{zhuang2020performance, li2021meta} that only send the backbone model $\Theta$ to the server, and keep the class embedding matrix on clients. The steps for collaborative training by server and clients are as follows:
\begin{enumerate}
    \item In the $t$-th communication round, the server sends the global model $\Theta^t_g$ to all client nodes.
    \item The $i$-th client updates the model $\Theta^t_{l(i)}$ at round $t$ based on $N_{l(i)}$ local data and local learned class embedding $W_{l(i)}$ with Cosface loss $\mathcal{L}_{cos}$, which is a $K_{l(i)}$-class classification problem.
    \item The local clients only send the backbone model $\Theta^t_{l(i)}$ to the server. The server will update the global model by taking a weighted average of them as follows:
    \begin{equation}
    \label{eq:theta}
        \Theta^{t+1}_{g} = \frac{1}{N}\sum_{i \in [C]} N_{l(i)} \cdot \Theta^t_{l(i)}, 
    \end{equation}
    where $N$ is the total number of training images across all client nodes.
    \item Last, the updated global model will then be transmitted to each client and steps $2-4$ are repeated until convergence.
\end{enumerate}
FedAvg can perform well on clients with IID-distributed data. However, for our face recognition setup, the identity distributions on each client are different. Just optimizing on local data with limited number of identities to obtain $\Theta^t_{l(i)}$ could harm the original performance of the pre-trained model (as shown in the experimental results). Furthermore, although $\Theta^t_{l(i)}$ can improve the personalized representation for these $K_{l(i)}$ identities, it will be continuously updated by the global model along the communication rounds, which cannot achieve optimal performance for the local users.

\subsection{FedFR: Joint Optimization Federated Framework}
To tackle the issues in FedAvg, we propose a joint optimization framework \textbf{FedFR}, which can effectively improve the generic face representation at the server with the use of globally shared data, and also optimize the personalized recognition performance simultaneously at local clients. We first provide an overview of FedFR, and the system architecture is also illustrated in Figure~\ref{fig:big}. Built upon the baseline FL pipeline, we introduce several novel techniques: \textbf{1)} We employ the globally shared dataset $D_g$ to better regularize the local model training, which could prevent the model from over-fitting on local identities. \textbf{2)} The Hard Negative Sampling strategy is introduced to select the most critical data from $D_g$ to significantly reduce the computation on local clients. \textbf{3)} The Contrastive Regularization is employed to control the drift of model parameters and better bridge the gap between global and local representations. \textbf{4)} To simultaneously optimize face representation for local clients, we propose the \textbf{Decoupled Feature Customization} module to transform the global representation for better fitting the local distributions. The corresponding margin-based binary classification loss $\mathcal{L}_{BCE}$ establishes a better local objective to supervise the learning of the decoupled branch. We elaborate each technique in details as follows.

\paragraph{Leveraging Globally Shared Data.} Some previous FL works on image classification~\cite{zhao2018federated, lin2020ensemble} has shown that leveraging globally shared dataset can better address the issue of heterogeneous clients. In the face recognition FL setup, the global dataset $D_g$ which was used for pre-training the server model can be naturally shared to all the local clients. We could further regularize the training of local clients by providing the class embedding matrix $\Phi_g$ of the shared $K_g$ identities. As shown in Figure~\ref{fig:big}, given the shared dataset $D_g$ on client $i$, the local client could build a more ``balanced objective'' by concatenating $\Phi^t_{l(i)}=\Phi^t_g$ with the local private embedding matrix $W_{l(i)}$ as a new learnable proxies and learn to classify $K_g + K_{l(i)}$ identities with $\mathcal{L}_{cos}$. Thus, our balanced Cosface loss would be formulated as:
\begin{equation}
    \mathcal{L}_{cos} = - \log \frac{e^{s(\cos\theta_{y}-m)}}{e^{s(\cos\theta_{y}-m)} + \sum_{j\neq y}^{K_g+K_{l(i)}}e^{s\cos\theta_{j}}},
\end{equation}
where the denominator is added with additional $K_g$ negative terms. For the end of each round $t$, beside sending the backbone $\Theta^t_{l(i)}$ back to server, the learned class embeddings $\Phi^t_{l(i)}$ related to $K_g$ global identities can also be sent back and updated by:
\begin{equation}
\label{eq:phi}
        \Phi^{t+1}_{g} = \frac{1}{N}\sum_{i \in [C]} N_{l(i)} \cdot \Phi^t_{l(i)} .
\end{equation}

\paragraph{Hard Negative Sampling Strategy.} Jointly training with $D_g$ can prevent model from over-fitting on local data. However, the large number of public data will also increase the computation burden on local clients, which will enlarge the training time and degrade the communication efficiency between server and clients. To obtain a better trade-off, we propose a hard negative (HN) sampling strategy to only choose a subset $D_{HN}$ from $D_g$, which is critical for learning with $D_{l(i)}$. The proposed technique is described as follows.

At the start of each communication round $t$ on local client $i$, we first forward the global and local data to $\Theta^t_g$ to generate their features. Then we can calculate the pair-wise cosine similarity between them. To make the training more efficient but at the same time maintain the performance, we only sample the ``hard'' global data for model learning, which is with similarity larger than threshold $t_{HN}$ to any of the local data. Intuitively, with larger $t_{HN}$, the less global data will be used for training. We decide the threshold by leveraging the inherent feature space of the pre-trained model. As mentioned above, the pre-trained model is trained with Cosface loss, where the similarity of each sample to its proxy should be larger than those to others by a margin $m$. Thus, if any negative pair with similarity larger than $t_{HN}=m$, they should be served as a hard negative pair. 

\paragraph{Contrastive Regularization on Local Clients.}
Inspired by the related work~\cite{li2021model}, which proposed a model-contrastive loss on the local training to prevent local model from deviating too much from the global model, we also apply the similar regularization on our face recognition task. Namely, we aim to decrease the distance between the face representation learned by the local model at time $t$ ($f = \Theta^t_{l(i)}(x) $) and the one learned by the global model ($f_{glob} = \Theta^t_g(x)$), and increase the distance between the face representation learned by the local model at time $t$ ($f = \Theta^t_{l(i)}(x)$) and time $t-1$ ($f_{prev} = \Theta^{t-1}_{l(i)}$). Thus, the local contrastive loss term $\mathcal{L}_{con}$ is defined as
\begin{equation}
    \small \mathcal{L}_{con}  = -\log{\frac{\exp(\textrm{sim}(f, f_{glob})/\tau)}{\exp(\textrm{sim}(f, f_{glob})/\tau) + \exp(\textrm{sim}(f, f_{prev})/\tau)}}, 
\end{equation}
where ``$\textrm{sim}(\cdot,\cdot)$'' measures the cosine similarity between face features, and $\tau$ denotes a temperature hyperparameter.

\paragraph{Decoupled Feature Customization.}
With the contrastive regularization, the local model can avoid over-parameterizing for the local objective and continuously improve the generic face representation. However, it will go against the goal which we aim to simultaneously obtain a personalized model to improve the local user experience. Thus, as shown in Figure~\ref{fig:big}, we propose a novel Decoupled Feature Customization (DFC) module to resolve this seemly contradicting scenario. In order not to influence the feature $f$ for generic representation, inspired by~\cite{wang2020unified}, we adopt a transformation $\Pi(f)$ with a fully-connected layer to map it to a client-specific feature space, which can recognize the $K_{l(i)}$ identities well. To achieve this goal, there should be a local objective for optimization. Inspired by~\cite{wen2021sphereface2}, we propose to adopt the binary classification on each local identity for the personalized purpose. Given the transformed feature $f^{\prime}=\Pi(f)$, we will feed it into $K_{l(i)}$ binary classification branches (which the total trainable weight vectors are denoted as $\Omega_{l(i)}$). The $k$-th module contains learnable parameters which only target on classifying the positive samples from the $k$-th class and the negative samples from ``any other'' classes. Formally, we follow the loss in the related work that used margin-based binary cross-entropy ($\mathcal{L}_{BCE}$) to supervise our personalized branch:
\begin{equation}\label{sphereface2_final}
\footnotesize
\begin{aligned}
    \mathcal{L}_{BCE} &=\frac{\lambda}{s^\prime}\cdot\log\bigg(1+\exp\big(-s^\prime\cdot (g(\cos\theta_k)-m^\prime)-b\big)\bigg)\\[-0.2mm]
    &+\frac{1-\lambda}{s^\prime}\cdot\sum_{j\neq k}\log\bigg(1+\exp \big(s^\prime\cdot (g(\cos\theta_j)+m^\prime)+b\big)\bigg),
\end{aligned}
\end{equation}
where $\cos\theta_j$ is the cosine similarity of transformed input feature $f^\prime$ and the $j$-th weight vector $\Omega_{l(i),j}$ in the binary classification, $b$ is the learned bias, and the function $g(z)=2((z+1)^{t^\prime} /2 )- 1$ is used to increase the empirical dynamic range of cosine similarity. The notations $\lambda,s^\prime$ and $m^\prime$ all follow those in the related work, which are the balanced factor, scaling constant and cosine margin.

It is worth mentioning that although there are only $K_{l(i)}$ binary classification branches, not only the local data but the global data can be used to optimize our DFC module because each branch only needs to recognize ``whether it is the $k$-th identity or not''. This objective just well-fits our personalized goal that given an unseen query image, a well-performed local face recognition system should quickly determine whether it is the registered identity or not.  

\paragraph{Learning Pipeline}
Our overall learning framework is based on FedAvg, where there will be $T$ communication rounds and in each round, the local clients will update the model for $E$ epochs. In the local client training, the model will be optimized in an end-to-end manner with the total objective $\mathcal{L}_{total}$, which is formulated as:
\begin{equation}
    \mathcal{L}_{total} = \alpha_1\mathcal{L}_{cos}+\alpha_2\mathcal{L}_{con}+\alpha_3\mathcal{L}_{BCE},
\end{equation}
where all the modules $\Theta^t_{l(i)},\Phi^t_{l(i)},W^t_{l(i)},\Pi^t_{l(i)},\Omega^{t}_{l(i)}$ and bias $b$ would be updated. However, only the $\Theta^t_{l(i)}$ and $\Phi^t_{l(i)}$ will be sent back for globally averaged with Equation~\ref{eq:theta} and~\ref{eq:phi}. In the testing phase, $\Theta_g$ is used for generic evaluation and $[\Theta_{l(i)},\Pi_{l(i)}]$ is used for personalized evaluation.
\vspace{-2mm}
\section{Experiments}
\begin{table*}[t!]
    \centering
    \ra{1.1}
    \label{tab:ablation}
    \scalebox{0.92}{
    \begin{tabular}{p{1.5cm}|ccc||cc|cc|cc|cc}
    \hline
    \multirow{3}{*}{\textbf{~~Setup}} & \multicolumn{3}{c||}{\textbf{Modules}}  & \multicolumn{4}{c|}{\textbf{Generic Evaluation} (IJB-C)}  & \multicolumn{4}{c}{\textbf{Personalized Evaluation}} \\
    \cline{2-12}
      &\multirow{2}{*}{\small \makecell{HN. sampled\\ Global data}}& \multirow{2}{*}{\small Contrastive}&\multirow{2}{*}{\small \makecell{DFC. \\ Branch}} &\multicolumn{2}{c|}{\small 1:1 TAR @ FAR} &\multicolumn{2}{c|}{\small 1:N TPIR @ FPIR} & \multicolumn{2}{c|}{\small 1:1 TAR @ FAR} & \multicolumn{2}{c}{\small 1:N TPIR @ FPIR} \\
       &&  & & 1e-5 & 1e-4 &  1e-2 & 1e-1 & 1e-6 & 1e-5 &1e-5 & 1e-4 \\
    \hline \hline
    \multicolumn{4}{l||}{Centrally trained on 6k IDs (pre-training)} &76.42&84.58 & 72.06& 80.30 &56.28&72.50& 71.73 & 82.33\\
    \hline
    \multirow{4}{*}{\makecell{Federated \\Learning \\ on 4k IDs}}&  \xmark &\xmark&\xmark&73.79&83.71 &67.59 &78.53  &67.33&85.70& 82.77 & 92.27\\
      &\cmark &\xmark&\xmark&76.79&84.64 & 72.76& 80.76 &81.75&91.91 & 91.97 & 96.09 \\
     & \cmark &\cmark&\xmark&77.41&85.17& 73.60 & 81.25 & 77.77 & 89.57 & 89.58 & 94.60\\
      &\cmark &\cmark&\cmark&\textbf{77.60}&\textbf{85.21}& \textbf{73.60} & \textbf{81.27} & \textbf{88.32} & \textbf{95.46} &\textbf{95.17} &\textbf{97.94}\\
    \hline\hline
    \multicolumn{4}{l||}{Centrally trained on 10k IDs}&77.56&85.99 & 73.30 &82.14 &93.72&97.39& 98.58& 99.40\\
    \hline
    \end{tabular}}
    \caption{Ablation Studies. We conduct FL experiments with 40 clients; each client contains 100 identities. (results are in \%)}
    \vspace{-2mm}
\end{table*}   
\subsection{Experimental Setup}
\paragraph{Dataset}
We use the MS-Celeb-1M~\cite{guo2016ms} as the training dataset. To avoid the long-tail distribution, we manually select 10k identities from the dataset where each identity contains 100 face images. Within the selected subset, 6000 identities ($K_g$) are used for pre-training the global model, and the other 4000 identities are equally distributed into local clients. For each identity in each local client, we use 60 images for local training, and 40 images for personalized model evaluation, respectively. Besides MS-Celeb-1M, IJB-C~\cite{maze2018iarpa} dataset which contains 3531 identities with diverse appearance is used for evaluating the generic model performance. The selected list for FL training will be released for fair comparison in the future.   

\paragraph{Evaluation Metrics}
For the generic model evaluation, we strictly follow the IJB-C evaluation protocol, which is commonly used in the face recognition community. We report the true acceptance rates (TAR) at different false acceptance rates (FAR) for 1:1 verification protocol, and true positive identification rates (TPIR) at different false positive identification rates (FPIR) for 1:N identification protocol.

Regarding the personalized model evaluation, we carefully build up the metrics and protocols as we are the first to investigate the personalized face recognition setup. The evaluation is supposed to only focus on the face recognition user experience of the registered identities on each local client. Therefore, we establish two evaluation protocols to better measure the client-specific performance: 1) Firstly, similar to the 1:1 verification protocol in IJB-C, we establish a list of positive pairs and negative pairs for evaluation. In each client, we formulate genuine matches from local identities and build up imposter matches by pairing one local identity with a random identity from other clients. For the 40 local clients scenario where each client is registered with 100 identities, there are $7.8$k positive pairs and about $630$ million negative pairs in one client. We average the true acceptance rates (TAR) across all clients as the final personalized verification performance. 2) Secondly, we build up an 1:N identification protocol to estimate the login experience on a local client (device). Intuitively, the registered images from one local identity are combined to form its gallery feature. And the testing images from all clients are taken as the probe features. For the 40 local clients scenario, there are 100 gallery features and $160$k probe features in one client. Similarly, we average the true positive identification rates (TPIR) across all clients as the final personalized identification performance. 
\vspace{-2mm}
\paragraph{Implementation Details}
For the backbone face model, we adopt the same 64-layer CNN architecture from~\cite{liu2017sphereface,wang2018cosface},
which outputs a 512-dimensional feature vector. The image preprocessing techniques are the same as~\cite{deng2019arcface}, where the image is cropped to size $112\times112$ and the pixel value is normalized to $[-1,1]$. To simplify our network training, all hyper-parameters in $\mathcal{L}_{cos}$, $\mathcal{L}_{con}$ and $\mathcal{L}_{BCE}$ are empirically set as the same ones in the related work, where $m$=$m^\prime$=$0.4$, $s$=$s^\prime$=$30$, $\tau$=$0.5$, $\lambda$=$0.7$ and $t^\prime$=$3$. For $\mathcal{L}_{total}$, the $\alpha_1$, $\alpha_2$ and $\alpha_3$ are empirically set as $1$, $5$ and $10$. We adopt SGD optimizer with weight decay $5\times 10^{-4}$ and learning rate $0.001$. For the FL setup, we conduct $T$=$30$ communication rounds and in each round the local clients  conduct $E$=$4$ epochs.

\subsection{Ablation Studies}
\paragraph{Effectiveness of each modules}
To validate the effectiveness of each proposed module, we report the ablation studies in Table.~\ref{tab:ablation}. The experiments are conducted with one central server and 40 clients, where each client contains 100 identities. The performance is evaluated both on the generic and personalized benchmark. If it is under the FL setup, the global model $\Theta_g$ will be used to test on the generic evaluation and each local model $\Theta_{l(i)}$ will be tested on personalized data, where the shown scores are the average over all clients. Notes that for the 1:N identification in personalized evaluation, we average the feature of training images based on their identities as the gallery features in that client.

The first row is the performance of pre-trained model trained on public data with 6k classes, which is the target model that needs to be improved. Start from $2^{nd}$ to $5^{th}$ row, the FL setup is employed where 4k augmented IDs are added but with privacy constraints. And for the last row, it is the ideal situation that we can centrally optimize the model with data of 10k IDs. We can see that in the second row, our baseline method which directly optimizes the model with local data and perform FedAvg on the server cannot perform well. The performance is even worse than the pre-trained one owing to the over-fitting on local data. Leveraging the public data is a solution, but it may suffer from long training time and large computation overhead. With our proposed Hard Negative sampling strategy where only a subset of global data serving as negative pairs to the local data, in the $3^{rd}$ row, not only the generic representation but also the personalized evaluation can be boosted. Contrastive regularization is designed for regularizing the local model from training towards the undesired local minimum. We can see that in the $4^{th}$ row, the performance improves greatly on generic evaluation. However, under the same feature space parameterized by $\Theta$, a more generalized representation will harm the performance for recognizing specific identities on clients. Thus, in the $5^{th}$ row, which is our final proposed FedFR architecture with the DFC branch, we decouple the feature from the original feature space to a new one with a transformation $\Pi_{l(i)}$ , and optimize this space with binary cross-entropy loss tailored for the personalization. We can see that with $\Theta_g$ for generic representation and $[\Theta_{l(i)},\Pi_{l(i)}]$ for personalized evaluation, both of them can achieve superior results.

\begin{figure}[t]
	\centering
    \includegraphics[width=0.82\linewidth]{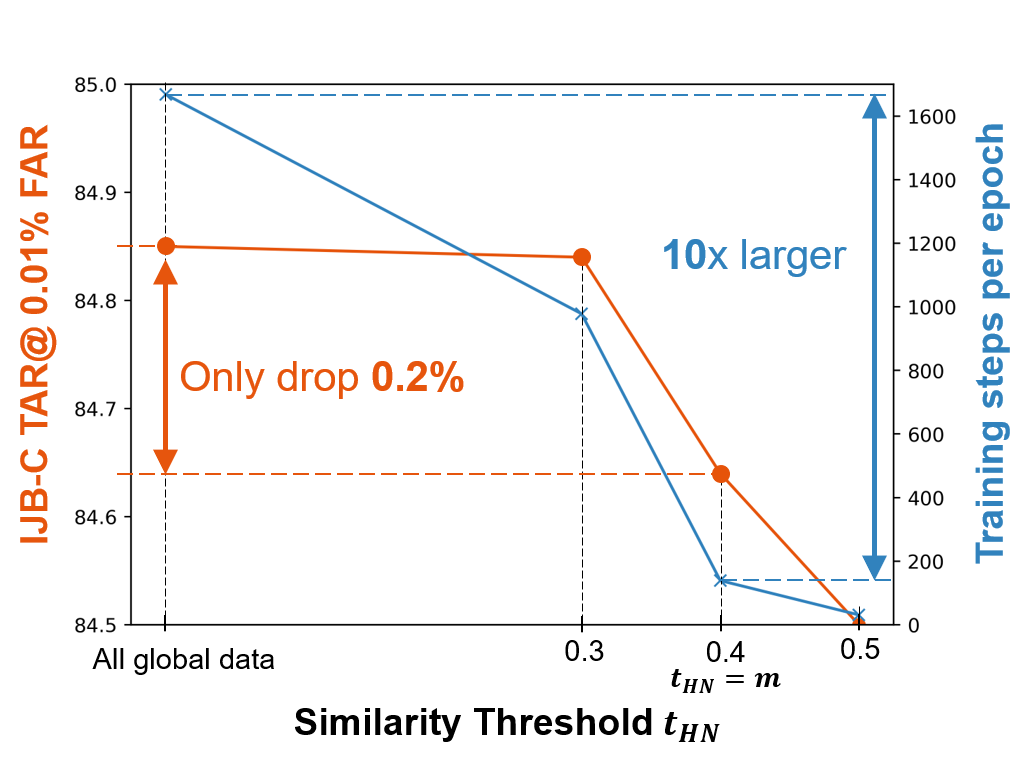}
    \vspace{-3mm}
    \caption{The generic model performance and the model training efficiency under different Hard Negative thresholds.}
    \label{fig:HN}
    \vspace{-3mm}
\end{figure}

\paragraph{Analysis of the $t_{HN}$ in Hard Negative Sampling}
In our experiments, we choose $t_{HN}$ equals to the margin $m$=$0.4$ in Cosface used in pre-training the model. To validate the effectiveness, as shown in Figure~\ref{fig:HN}, we demonstrate the global performance on IJB-C and its training efficiency under different hard negative thresholds with 100 IDs per client. The training efficiency is measured in terms of the training steps per epoch. We can see that with $t_{HN}$=$0.4$, the number of sampled global data can be largely reduced by $\mathbf{10}$ times but with only $0.2\%$ drop of the global performance, which is the best trade-off configuration in our experiments.

\vspace{-3mm}
\subsection{Comparison with FedFace}
\begin{figure}[t]
	\centering
    \includegraphics[width=0.85\linewidth]{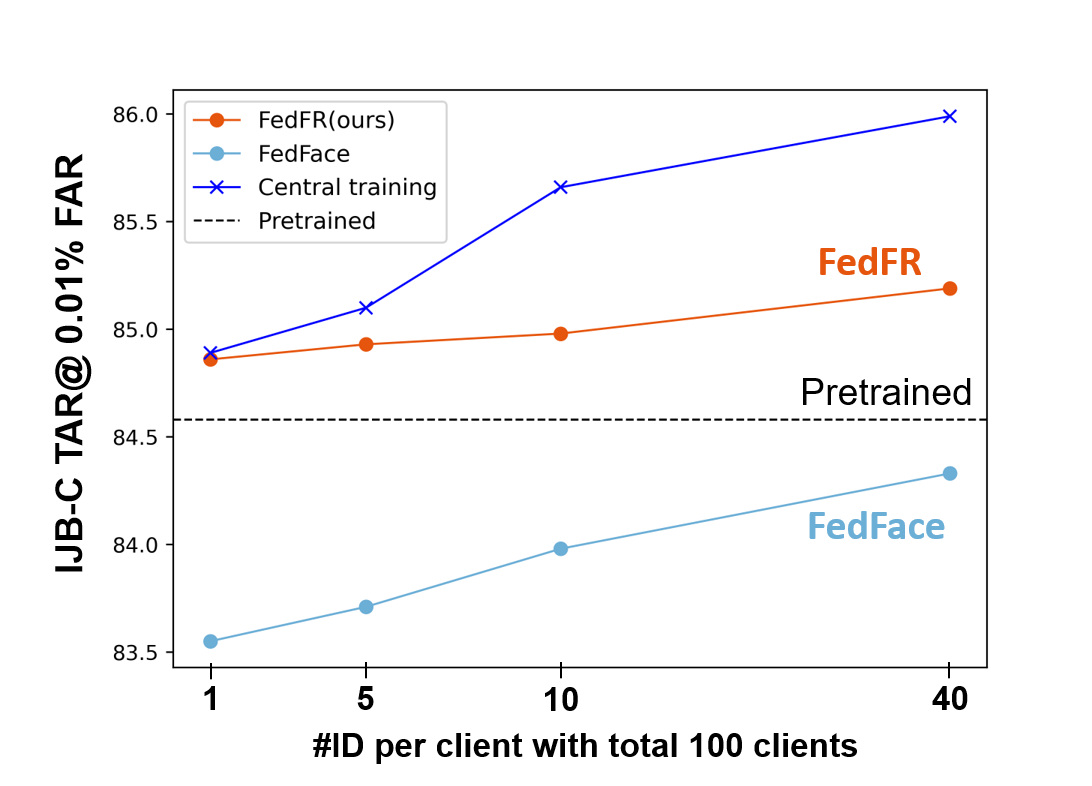}
    \vspace{-3mm}
    \caption{Generic model performance compared to FedFace. We fix the number of clients to 100 and conduct 4 scenarios of different \#IDs in one client.}
    \label{fig:fedface}
\end{figure}

\vspace{-1mm}
To compare the results with FedFace~\cite{aggarwal2021fedface}, as shown in Figure~\ref{fig:fedface}, we construct the FL setting with diverse identities per client under total 100 clients, which is from 40 to 1. We demonstrate the results of the pre-trained model, ideal central training (upper bound), FedFace and our proposed FedFR. Because FedFace cannot be adopted on multiple IDs in a client and their FL dataset is not released, we re-implement their method on our setting that uses Cosface loss as the local objective if the number of ID is larger than 1, and also apply spreadout regularizer at the server side to separate the class proxies from clients. From the comparison on the generic model performance, FedFace could easily over-fit on local dataset and performs inferior to the pre-trained model in these scenarios. In contrast, our proposed FedFR can still improve the generic face representation under the most challenging scenario where there is only one identity in the client.
\vspace{-3mm}
\subsection{Comparison with Personalized FL Methods}
\vspace{-1mm}
To validate the effectiveness of our Decoupled Feature Customization (DFC) module, we compare with the latest personalized FL method~\cite{yu2020salvaging}, which is a two-stage local adaptation approach. For fair comparison, we re-implement the ``Fine-tune'' and ``KD'' local adaptation methods, which were shown to be effective in image classification tasks, in our face recognition setup. In the first stage, the server and clients collaboratively learn to obtain a great generic model, where we use the proposed hard negative sampling strategy and the contrastive regularization in the experiments. Then, in the second stage, each client separately optimizes its local model for personalization. For the ``Fine-tune'' method, we directly optimize each model with Cosface loss with the local and sampled global data. For the ``KD'' method, it is with a Knowledge Distillation technique that besides the original Cosface loss, we also supervise the output logits of local model (student) by the logits generated from original global model (teacher) with KL-Divergence loss. As illustrated in Table.~\ref{tab:local}, our proposed one-stage personalization method can outperform the two local adaptation strategies. In addition, we also conduct a variant of our method, which is also a decoupled branch but adopts a Cosface loss with multi-class classification for supervision. It is clearly verified that the proposed binary classification objective better fits the need for the personalized face recognition on clients.

\begin{table}[t]
    \centering
    \ra{1.0}
    \label{tab:local}
    \vspace{-2mm}
    \scalebox{0.83}{
    \begin{tabular}{c|c|cc|cc}
    \hline
    \multirow{3}{*}{\textbf{Method}} &\multirow{3}{*}{\textbf{Modules}}& \multicolumn{4}{c}{\textbf{Personalized Evaluation}} \\
    \cline{3-6}
       & & \multicolumn{2}{c|}{\small 1:1 TAR @ FAR} & \multicolumn{2}{c}{\small 1:N TPIR @ FPIR} \\
       & & 1e-6 & 1e-5 &1e-5 & 1e-4 \\
    \hline \hline
    \multirow{2}{*}{\makecell{Yu et al.\\ 2020}}& Fine-tune  &73.81 &86.21 &88.37  & 93.90\\
     &KD & 75.82& 87.65 &89.50 & 94.67 \\
    \hline
    \multirow{2}{*}{\makecell{\textbf{Ours} \\(w/ branch)}} & Cosface  & 82.93 & 91.88 &90.67 & 95.59 \\
     & \textbf{BCE} & \textbf{88.32} & \textbf{95.46} & \textbf{95.17} & \textbf{97.94} \\
    \hline
    \end{tabular}}
    \caption{Comparison of other personalized techniques. It is conducted on 40 clients with 100 IDs per each. }    
    \vspace{-4mm}
\end{table}

\vspace{-3mm}
\section{Conclusion}
\vspace{-1mm}
In this paper, we address the face recognition model training under the practical federated learning setting, where each client is initialized with the pre-trained model. We propose a novel joint optimization framework FedFR, which can improve the generic face representation of the global model and at the same time enhance the personalized user experience. While the proposed hard negative sampling and contrastive regularization can efficiently bridge the gap between global and local training, the Decoupled Feature Customization (DFC) module is another novel component to enable concurrent optimization of the personalized face recognition model. The effectiveness of the proposed solution is verified on several challenging generic and personalized face recognition benchmarks. We hope that the work and the release of the personalized FR benchmark can facilitate the future research on the federated learning for face recognition.
\section{Acknowledgments}
This research was supported in part by the Ministry of Science and Technology of Taiwan (MOST 109-2218-E-002 -026), National Taiwan University (NTU-108L104039), Intel Corporation, Delta Electronics and Compal Electronics. 
\bibliographystyle{aaai} 	
\bibliography{main}

\begin{thebibliography}{38}
\providecommand{\natexlab}[1]{#1}

\bibitem[{Aggarwal, Zhou, and Jain(2021)}]{aggarwal2021fedface}
Aggarwal, D.; Zhou, J.; and Jain, A.~K. 2021.
\newblock FedFace: Collaborative Learning of Face Recognition Model.
\newblock \emph{arXiv preprint arXiv:2104.03008}.

\bibitem[{Bonawitz et~al.(2019)Bonawitz, Eichner, Grieskamp, Huba, Ingerman,
  Ivanov, Kiddon, Kone{\v{c}}n{\`y}, Mazzocchi, McMahan
  et~al.}]{bonawitz2019towards}
Bonawitz, K.; Eichner, H.; Grieskamp, W.; Huba, D.; Ingerman, A.; Ivanov, V.;
  Kiddon, C.; Kone{\v{c}}n{\`y}, J.; Mazzocchi, S.; McMahan, H.~B.; et~al.
  2019.
\newblock Towards federated learning at scale: System design.
\newblock \emph{arXiv preprint arXiv:1902.01046}.

\bibitem[{Cao et~al.(2018)Cao, Shen, Xie, Parkhi, and
  Zisserman}]{cao2018vggface2}
Cao, Q.; Shen, L.; Xie, W.; Parkhi, O.~M.; and Zisserman, A. 2018.
\newblock Vggface2: A dataset for recognising faces across pose and age.
\newblock In \emph{2018 13th IEEE international conference on automatic face \&
  gesture recognition (FG 2018)}.

\bibitem[{Chen et~al.(2018)Chen, Luo, Dong, Li, and He}]{chen2018federated}
Chen, F.; Luo, M.; Dong, Z.; Li, Z.; and He, X. 2018.
\newblock Federated meta-learning with fast convergence and efficient
  communication.
\newblock \emph{arXiv preprint arXiv:1802.07876}.

\bibitem[{Chen and Chao(2021)}]{chen2021bridging}
Chen, H.-Y.; and Chao, W.-L. 2021.
\newblock On Bridging Generic and Personalized Federated Learning.
\newblock \emph{arXiv preprint arXiv:2107.00778}.

\bibitem[{Deng et~al.(2019)Deng, Guo, Xue, and Zafeiriou}]{deng2019arcface}
Deng, J.; Guo, J.; Xue, N.; and Zafeiriou, S. 2019.
\newblock Arcface: Additive angular margin loss for deep face recognition.
\newblock In \emph{CVPR}.

\bibitem[{Duong et~al.(2020)Duong, Truong, Luu, Quach, Bui, and
  Roy}]{duong2020vec2face}
Duong, C.~N.; Truong, T.-D.; Luu, K.; Quach, K.~G.; Bui, H.; and Roy, K. 2020.
\newblock Vec2Face: Unveil Human Faces From Their Blackbox Features in Face
  Recognition.
\newblock In \emph{CVPR}.

\bibitem[{Fallah, Mokhtari, and Ozdaglar(2020)}]{fallah2020personalized}
Fallah, A.; Mokhtari, A.; and Ozdaglar, A. 2020.
\newblock Personalized federated learning: A meta-learning approach.
\newblock \emph{arXiv preprint arXiv:2002.07948}.

\bibitem[{Guo et~al.(2016)Guo, Zhang, Hu, He, and Gao}]{guo2016ms}
Guo, Y.; Zhang, L.; Hu, Y.; He, X.; and Gao, J. 2016.
\newblock Ms-celeb-1m: A dataset and benchmark for large-scale face
  recognition.
\newblock In \emph{ECCV}.

\bibitem[{Haddadpour and Mahdavi(2019)}]{haddadpour2019convergence}
Haddadpour, F.; and Mahdavi, M. 2019.
\newblock On the convergence of local descent methods in federated learning.
\newblock \emph{arXiv preprint arXiv:1910.14425}.

\bibitem[{He et~al.(2016)He, Zhang, Ren, and Sun}]{he2016deep}
He, K.; Zhang, X.; Ren, S.; and Sun, J. 2016.
\newblock Deep residual learning for image recognition.
\newblock In \emph{CVPR}.

\bibitem[{Kairouz et~al.(2019)Kairouz, McMahan, Avent, Bellet, Bennis, Bhagoji,
  Bonawitz, Charles, Cormode, Cummings et~al.}]{kairouz2019advances}
Kairouz, P.; McMahan, H.~B.; Avent, B.; Bellet, A.; Bennis, M.; Bhagoji, A.~N.;
  Bonawitz, K.; Charles, Z.; Cormode, G.; Cummings, R.; et~al. 2019.
\newblock Advances and open problems in federated learning.
\newblock \emph{arXiv preprint arXiv:1912.04977}.

\bibitem[{Khaled, Mishchenko, and Richt{\'a}rik(2020)}]{khaled2020tighter}
Khaled, A.; Mishchenko, K.; and Richt{\'a}rik, P. 2020.
\newblock Tighter theory for local SGD on identical and heterogeneous data.
\newblock In \emph{International Conference on Artificial Intelligence and
  Statistics}. PMLR.

\bibitem[{Kone{\v{c}}n{\`y} et~al.(2016)Kone{\v{c}}n{\`y}, McMahan, Yu,
  Richt{\'a}rik, Suresh, and Bacon}]{konevcny2016federated}
Kone{\v{c}}n{\`y}, J.; McMahan, H.~B.; Yu, F.~X.; Richt{\'a}rik, P.; Suresh,
  A.~T.; and Bacon, D. 2016.
\newblock Federated learning: Strategies for improving communication
  efficiency.
\newblock \emph{arXiv preprint arXiv:1610.05492}.

\bibitem[{Kulkarni, Kulkarni, and Pant(2020)}]{kulkarni2020survey}
Kulkarni, V.; Kulkarni, M.; and Pant, A. 2020.
\newblock Survey of personalization techniques for federated learning.
\newblock In \emph{2020 Fourth World Conference on Smart Trends in Systems,
  Security and Sustainability (WorldS4)}.

\bibitem[{Li et~al.(2021{\natexlab{a}})Li, Niu, Jiang, Zuo, and
  Yang}]{li2021meta}
Li, C.; Niu, D.; Jiang, B.; Zuo, X.; and Yang, J. 2021{\natexlab{a}}.
\newblock Meta-HAR: Federated Representation Learning for Human Activity
  Recognition.
\newblock \emph{arXiv preprint arXiv:2106.00615}.

\bibitem[{Li, He, and Song(2021)}]{li2021model}
Li, Q.; He, B.; and Song, D. 2021.
\newblock Model-Contrastive Federated Learning.
\newblock In \emph{CVPR}.

\bibitem[{Li et~al.(2019{\natexlab{a}})Li, Wen, Wu, Hu, Wang, Li, Liu, and
  He}]{li2019survey}
Li, Q.; Wen, Z.; Wu, Z.; Hu, S.; Wang, N.; Li, Y.; Liu, X.; and He, B.
  2019{\natexlab{a}}.
\newblock A survey on federated learning systems: vision, hype and reality for
  data privacy and protection.
\newblock \emph{arXiv preprint arXiv:1907.09693}.

\bibitem[{Li et~al.(2019{\natexlab{b}})Li, Huang, Yang, Wang, and
  Zhang}]{li2019convergence}
Li, X.; Huang, K.; Yang, W.; Wang, S.; and Zhang, Z. 2019{\natexlab{b}}.
\newblock On the convergence of fedavg on non-iid data.
\newblock \emph{arXiv preprint arXiv:1907.02189}.

\bibitem[{Li et~al.(2021{\natexlab{b}})Li, Jiang, Zhang, Kamp, and
  Dou}]{li2021fedbn}
Li, X.; Jiang, M.; Zhang, X.; Kamp, M.; and Dou, Q. 2021{\natexlab{b}}.
\newblock Fedbn: Federated learning on non-iid features via local batch
  normalization.
\newblock \emph{arXiv preprint arXiv:2102.07623}.

\bibitem[{Liang et~al.(2020)Liang, Liu, Ziyin, Allen, Auerbach, Brent,
  Salakhutdinov, and Morency}]{liang2020think}
Liang, P.~P.; Liu, T.; Ziyin, L.; Allen, N.~B.; Auerbach, R.~P.; Brent, D.;
  Salakhutdinov, R.; and Morency, L.-P. 2020.
\newblock Think locally, act globally: Federated learning with local and global
  representations.
\newblock \emph{arXiv preprint arXiv:2001.01523}.

\bibitem[{Lin et~al.(2020)Lin, Kong, Stich, and Jaggi}]{lin2020ensemble}
Lin, T.; Kong, L.; Stich, S.~U.; and Jaggi, M. 2020.
\newblock Ensemble distillation for robust model fusion in federated learning.
\newblock \emph{arXiv preprint arXiv:2006.07242}.

\bibitem[{Liu et~al.(2017)Liu, Wen, Yu, Li, Raj, and Song}]{liu2017sphereface}
Liu, W.; Wen, Y.; Yu, Z.; Li, M.; Raj, B.; and Song, L. 2017.
\newblock Sphereface: Deep hypersphere embedding for face recognition.
\newblock In \emph{CVPR}.

\bibitem[{Maze et~al.(2018)Maze, Adams, Duncan, Kalka, Miller, Otto, Jain,
  Niggel, Anderson, Cheney et~al.}]{maze2018iarpa}
Maze, B.; Adams, J.; Duncan, J.~A.; Kalka, N.; Miller, T.; Otto, C.; Jain,
  A.~K.; Niggel, W.~T.; Anderson, J.; Cheney, J.; et~al. 2018.
\newblock Iarpa janus benchmark-c: Face dataset and protocol.
\newblock In \emph{2018 International Conference on Biometrics (ICB)}.

\bibitem[{McMahan et~al.(2017)McMahan, Moore, Ramage, Hampson, and
  y~Arcas}]{mcmahan2017communication}
McMahan, B.; Moore, E.; Ramage, D.; Hampson, S.; and y~Arcas, B.~A. 2017.
\newblock Communication-efficient learning of deep networks from decentralized
  data.
\newblock In \emph{Artificial intelligence and statistics}. PMLR.

\bibitem[{Meng et~al.(2022)Meng, Zhou, Ren, Feng, Liu, and
  Lin}]{meng2022improving}
Meng, Q.; Zhou, F.; Ren, H.; Feng, T.; Liu, G.; and Lin, Y. 2022.
\newblock Improving Federated Learning Face Recognition via Privacy-Agnostic
  Clusters.
\newblock \emph{arXiv preprint arXiv:2201.12467}.

\bibitem[{Schroff, Kalenichenko, and Philbin(2015)}]{schroff2015facenet}
Schroff, F.; Kalenichenko, D.; and Philbin, J. 2015.
\newblock Facenet: A unified embedding for face recognition and clustering.
\newblock In \emph{CVPR}.

\bibitem[{Sun et~al.(2020)Sun, Cheng, Zhang, Zhang, Zheng, Wang, and
  Wei}]{sun2020circle}
Sun, Y.; Cheng, C.; Zhang, Y.; Zhang, C.; Zheng, L.; Wang, Z.; and Wei, Y.
  2020.
\newblock Circle loss: A unified perspective of pair similarity optimization.
\newblock In \emph{CVPR}, 6398--6407.

\bibitem[{Wang et~al.(2020)Wang, Chang, Yang, Chen, and Lai}]{wang2020unified}
Wang, C.-Y.; Chang, Y.-L.; Yang, S.-T.; Chen, D.; and Lai, S.-H. 2020.
\newblock Unified representation learning for cross model compatibility.
\newblock \emph{arXiv preprint arXiv:2008.04821}.

\bibitem[{Wang et~al.(2018)Wang, Wang, Zhou, Ji, Gong, Zhou, Li, and
  Liu}]{wang2018cosface}
Wang, H.; Wang, Y.; Zhou, Z.; Ji, X.; Gong, D.; Zhou, J.; Li, Z.; and Liu, W.
  2018.
\newblock Cosface: Large margin cosine loss for deep face recognition.
\newblock In \emph{CVPR}.

\bibitem[{Wang et~al.(2021)Wang, Charles, Xu, Joshi, McMahan, Al-Shedivat,
  Andrew, Avestimehr, Daly, Data et~al.}]{wang2021field}
Wang, J.; Charles, Z.; Xu, Z.; Joshi, G.; McMahan, H.~B.; Al-Shedivat, M.;
  Andrew, G.; Avestimehr, S.; Daly, K.; Data, D.; et~al. 2021.
\newblock A Field Guide to Federated Optimization.
\newblock \emph{arXiv preprint arXiv:2107.06917}.

\bibitem[{Wen et~al.(2021)Wen, Liu, Weller, Raj, and
  Singh}]{wen2021sphereface2}
Wen, Y.; Liu, W.; Weller, A.; Raj, B.; and Singh, R. 2021.
\newblock SphereFace2: Binary Classification is All You Need for Deep Face
  Recognition.
\newblock \emph{arXiv preprint arXiv:2108.01513}.

\bibitem[{Yu et~al.(2020)Yu, Rawat, Menon, and Kumar}]{yu2020federated}
Yu, F.; Rawat, A.~S.; Menon, A.; and Kumar, S. 2020.
\newblock Federated learning with only positive labels.
\newblock In \emph{ICML}.

\bibitem[{Yu, Bagdasaryan, and Shmatikov(2020)}]{yu2020salvaging}
Yu, T.; Bagdasaryan, E.; and Shmatikov, V. 2020.
\newblock Salvaging federated learning by local adaptation.
\newblock \emph{arXiv preprint arXiv:2002.04758}.

\bibitem[{Zhang and Yang(2017)}]{zhang2017survey}
Zhang, Y.; and Yang, Q. 2017.
\newblock A survey on multi-task learning.
\newblock \emph{arXiv preprint arXiv:1707.08114}.

\bibitem[{Zhao et~al.(2018)Zhao, Li, Lai, Suda, Civin, and
  Chandra}]{zhao2018federated}
Zhao, Y.; Li, M.; Lai, L.; Suda, N.; Civin, D.; and Chandra, V. 2018.
\newblock Federated learning with non-iid data.
\newblock \emph{arXiv preprint arXiv:1806.00582}.

\bibitem[{Zhu et~al.(2021)Zhu, Huang, Deng, Ye, Huang, Chen, Zhu, Yang, Lu, Du
  et~al.}]{zhu2021webface260m}
Zhu, Z.; Huang, G.; Deng, J.; Ye, Y.; Huang, J.; Chen, X.; Zhu, J.; Yang, T.;
  Lu, J.; Du, D.; et~al. 2021.
\newblock WebFace260M: A Benchmark Unveiling the Power of Million-Scale Deep
  Face Recognition.
\newblock In \emph{Proceedings of the IEEE/CVF Conference on Computer Vision
  and Pattern Recognition}, 10492--10502.

\bibitem[{Zhuang et~al.(2020)Zhuang, Wen, Zhang, Gan, Yin, Zhou, Zhang, and
  Yi}]{zhuang2020performance}
Zhuang, W.; Wen, Y.; Zhang, X.; Gan, X.; Yin, D.; Zhou, D.; Zhang, S.; and Yi,
  S. 2020.
\newblock Performance optimization of federated person re-identification via
  benchmark analysis.
\newblock In \emph{Proceedings of the 28th ACM International Conference on
  Multimedia}.

\end{thebibliography}

\section{Supplementary Materials}
\subsection{Learning Pipeline of FedFR}
Algorithm~\ref{alg:algorithm} demonstrates our whole FedFR pipeline in detail.  Given a central server and $C$ local clients, we will conduct $T$ communication rounds and in each round, the local client conduct $E$ training epoch. After local training, all the $\Theta^{t}_{l(i)}$ and $\Phi^{t}_{l(i)}$ will be sent to server and perform FedAvg to generate $\Theta^{t+1}_g$ and $\Phi^{t+1}_g$. To evaluate the global generic evaluation, the model $\Theta_g$ will be used and for each local personalized evaluation, we will concatenate the $[\Theta_{l(i)},\Pi_{l(i)}]$ as the personalized backbone to generate customized feature representation.
\begin{algorithm}[tb]
\caption{The FedFR framework}
\label{alg:algorithm}
\textbf{Input}: number of communication rounds $T$, number of clients $C$, number of local epochs $E$, hard negative threshold $t_{HN}$, hyper-parameter $\alpha_1$, $\alpha_2$, $\alpha_3$, learning rate $\mu$ \\
\textbf{Output}: Final global model $\Theta_g$, Final personalized models $\Pi_{l(i)}, \Theta_{l(i)}$ (i = 1, 2, ... C)
\begin{algorithmic}[1] 
\STATE \textbf{Server executes:}
\STATE initialize $\Theta^0_g$ and $\Phi^0_g$ with the pre-trained model
\FOR{$t = 0, 1, ..., T-1$}
\FOR{$i = 1, 2, ..., C$}
\STATE Send the global models $\Theta^t_g$ and $\Phi^t_g$ to client i
\STATE $\Pi_{l(i)}, \Theta^t_{l(i)}, \Phi^t_{l(i)}$ $\leftarrow$
\textbf{ClientTraining} (i, $\Theta^t_g$, $\Phi^t_g$)
\ENDFOR
\STATE $\Theta^{t+1}_{g} \leftarrow \frac{1}{N}\sum_{i \in [C]} N_{l(i)} \cdot \Theta^t_{l(i)}$ \\
\STATE $\Phi^{t+1}_{g} \leftarrow \frac{1}{N}\sum_{i \in [C]} N_{l(i)} \cdot \Phi^t_{l(i)}$
\ENDFOR
\STATE return $\Theta^T_{g}$, $\Phi^T_{l(i)}$, $\Pi^T_{l(i)}$ (i = 1, 2, ... C)
\STATE ~~~
\STATE \textbf{ClientTraining} (i, $\Theta^t$, $\Phi^t$):
\STATE $\Theta^t_{l(i)}, \Phi^t_{l(i)} \leftarrow \Theta^t, \Phi^t  $
\STATE $W^t_{l(i)}, \Pi^t_{l(i)}, \Omega^t_{l(i)} \leftarrow W^{t-1}_{l(i)}, \Pi^{t-1}_{l(i)}, \Omega^{t-1}_{l(i)}$
\STATE $D_{HN}^t \leftarrow$ select hard negative sets with threshold $t_{HN}$ from $D_g$
\FOR{epoch e = 1, 2, ..., E}
\FOR{each batch \textbf{B} = \{x, y\} of $D_{l(i)} \bigcup D^{t}_{HN}$}
\STATE $\mathcal{L}_{total} \leftarrow \alpha_{1} \mathcal{L}_{cos}(\Theta^t_{l(i)},\Phi^t_{l(i)},W^t_{l(i)},\textbf{B})+\alpha_{2} \mathcal{L}_{con}(\Theta^t_{l(i)},\Theta^{t-1}_{l(i)},\Theta^{t},\textbf{B})+\alpha_{3} \mathcal{L}_{BCE}(\Theta^t_{l(i)},\Pi^t_{l(i)},\Omega^t_{l(i)},\textbf{B})$
\STATE $\Theta^t_{l(i)} \leftarrow \Theta^t_{l(i)} - \mu \bigtriangledown_{\Theta^t_{l(i)}}\mathcal{L}$
\STATE $\Phi^t_{l(i)} \leftarrow \Phi^t_{l(i)} - \mu \bigtriangledown_{\Phi^t_{l(i)}}\mathcal{L}$
\STATE $\Pi^t_{l(i)} \leftarrow \Pi^t_{l(i)} - \mu \bigtriangledown_{\Pi^t_{l(i)}}\mathcal{L}$
\STATE $W^t_{l(i)} \leftarrow W^t_{l(i)} - \mu \bigtriangledown_{W^t_{l(i)}}\mathcal{L}$
\STATE $\Omega^t_{l(i)} \leftarrow \Omega^t_{l(i)} - \mu \bigtriangledown_{\Omega^t_{l(i)}}\mathcal{L}$
\ENDFOR
\ENDFOR
\STATE \textbf{return} $\Pi^t_{l(i)},\Theta^t_{l(i)},\Phi^t_{l(i)}$
\end{algorithmic}
\end{algorithm}

\subsection{Models for Personalized Evaluation}
For personalized face recognition, we want to validate that using $[\Theta_{l(i)},\Pi_{l(i)}]$ can obtain the best results compared to using $\Theta_g$ or $\Theta_{l(i)}$ as the model for personalized evaluation. Table~\ref{tab:cross} demonstrate the results on different models trained under our FedFR framework. The first row is the personalized result of global model $\Theta_g$. We can see that although it can perform well on generic evaluation (IJB-C), it is not suitable for user customization. The second row is the local backbone $\Theta_{l(i)}$ without the concatenation of $\Pi_{l(i)}$. We can see that the local feature can improve the performance, but with our Decoupled Feature Customization, the personalized feature ($[\Theta_{l(i)},\Pi_{l(i)}]$) can well-fit the local distribution and outperform all the others.

\begin{table}[t]
    \centering
    \ra{1.0}
    \vspace{-2mm}
    \label{tab:cross}
    \begin{tabular}{l|cc|cc}
    \hline
    \multirow{3}{*}{\textbf{Evaluated Model}} & \multicolumn{4}{c}{\textbf{Personalized Evaluation}} \\
    \cline{2-5}
        & \multicolumn{2}{c|}{\small 1:1 TAR @ FAR} & \multicolumn{2}{c}{\small 1:N TPIR @ FPIR} \\
       &  1e-6 & 1e-5 &1e-5 & 1e-4 \\
    \hline \hline
    $\Theta_g$ &70.95 & 84.40 &80.30  & 88.98\\
    $\Theta_{l(i)}$ & 81.46 & 92.17 & 91.15& 95.64\\
     $[\Theta_{l(i)},\Pi_{l(i)}]$(\textbf{Ours}) & \textbf{88.32} & \textbf{95.46} & \textbf{95.17} & \textbf{97.94}\\
    \hline
    \end{tabular}
    \caption{Choices of models on Personalized Evaluation of FedFR}
\end{table}   

\end{document}